\documentclass[lettersize,journal]{IEEEtran}

\IEEEoverridecommandlockouts                              %

\usepackage{fancyhdr}
\title{\LARGE \bf
Local positional graphs and attentive local features for a data and runtime-efficient hierarchical place recognition pipeline
}

\author{Fangming Yuan, Stefan Schubert, Peter Protzel and Peer Neubert%
\thanks{This work was partially supported by the German Federal Ministry for Economic Affairs and Climate Action.
F.~Y., S.~S. and P.~P. are with the Chemnitz University of Technology, Germany. Email: \{fangming.yuan, stefan.schubert, protzel\}@etit.tu-chemnitz.de. P.~N. is with the University of Koblenz, Germany. Email: neubert@uni-koblenz.de.}}%

\usepackage{graphicx}
\usepackage{amssymb}
\usepackage{amsmath}
\usepackage{float}
\usepackage{array}
\usepackage{color}
\usepackage[inline]{enumitem}
\usepackage[
protrusion=true,
activate={true,nocompatibility},
final,
tracking=true,
kerning=true,
spacing=true,
factor=1100]{microtype}%

\usepackage{multirow}
\usepackage{array}
\usepackage{makecell}

\usepackage{balance}

\usepackage{hyperref}
\usepackage[nameinlink,capitalize]{cleveref}
\crefname{section}{Section}{Sections}
\crefname{subsection}{Section}{Sections}
\crefname{enumi}{question}{questions}

\usepackage[textsize=tiny]{todonotes}

\begin{document}

\maketitle

\fancyfoot{}
\fancyhead[OL]{ 
    \footnotesize
    To appear in IEEE Robotics and Automation Letters (RA-L), 2024. ACCEPTED VERSION\\
    \tiny
    \copyright 2024 IEEE. Personal use of this material is permitted.  Permission from IEEE must be obtained for all other uses, in any current or future media, including reprinting/republishing this material for advertising or promotional purposes, creating new collective works, for resale or redistribution to servers or lists, or reuse of any copyrighted component of this work in other works.
}
\addtolength{\headheight}{\baselineskip}
\thispagestyle{fancy}
\pagestyle{empty}

\begin{abstract}  %

Large-scale applications of Visual Place Recognition (VPR) require computationally efficient approaches.
Further, a well-balanced combination of data-based and training-free approaches can decrease the required amount of training data and effort and can reduce the influence of distribution shifts between the training and application phases.
This paper proposes a runtime and data-efficient hierarchical VPR pipeline that extends existing approaches and presents novel ideas.
There are three main contributions:
First, we propose Local Positional Graphs (LPG), a training-free and runtime-efficient approach to encode spatial context information of local image features.
LPG can be combined with existing local feature detectors and descriptors and considerably improves the image-matching quality compared to existing techniques in our experiments.
Second, we present Attentive Local SPED (ATLAS), an extension of our previous local features approach with an attention module that improves the feature quality while maintaining high data efficiency.
The influence of the proposed modifications is evaluated in an extensive ablation study.
Third, we present a hierarchical pipeline that exploits hyperdimensional computing to use the same local features as holistic HDC-descriptors for fast candidate selection and for candidate reranking.
We combine all contributions in a runtime and data-efficient VPR pipeline that shows
benefits over the state-of-the-art method Patch-NetVLAD on a large collection of standard place recognition datasets with 15$\%$ better performance in VPR accuracy, 54$\times$ faster feature comparison speed, and 55$\times$ less descriptor storage occupancy, making our method promising for real-world high-performance large-scale VPR in changing environments. 
Code will be made available with publication of this paper.

\end{abstract}

\section{Introduction}\label{sec:intro}
Visual place recognition (VPR) is a crucial component of SLAM systems. It finds the most similar images in a database with the given query image for various use cases such as loop closure detection or (re-)localization.
The task of VPR is particularly challenging when the environmental condition between the database and the query set changes, e.g., from day to night or from summer to winter.

Convolutional neural networks (CNNs) based VPR methods can extract environmental condition- and viewpoint-robust image descriptors.
These methods extract the descriptor of an image to either a single holistic (global) feature for the whole image~\cite{SPED} or a set of local features for regions of interest in the image~\cite{D2Net}.
Among CNN-based VPR methods, learning-based attentive local feature methods provide superior VPR performance. 
Compared to methods that densely extract large amounts of local features~\cite{Patch-NetVLAD}~\cite{CTU-ICRA-Paper}, the attentive methods extract sparsely distributed and highly representative local features. 
	\begin{figure}[t]   
	    \centering 
  		\includegraphics[width=0.9\linewidth]{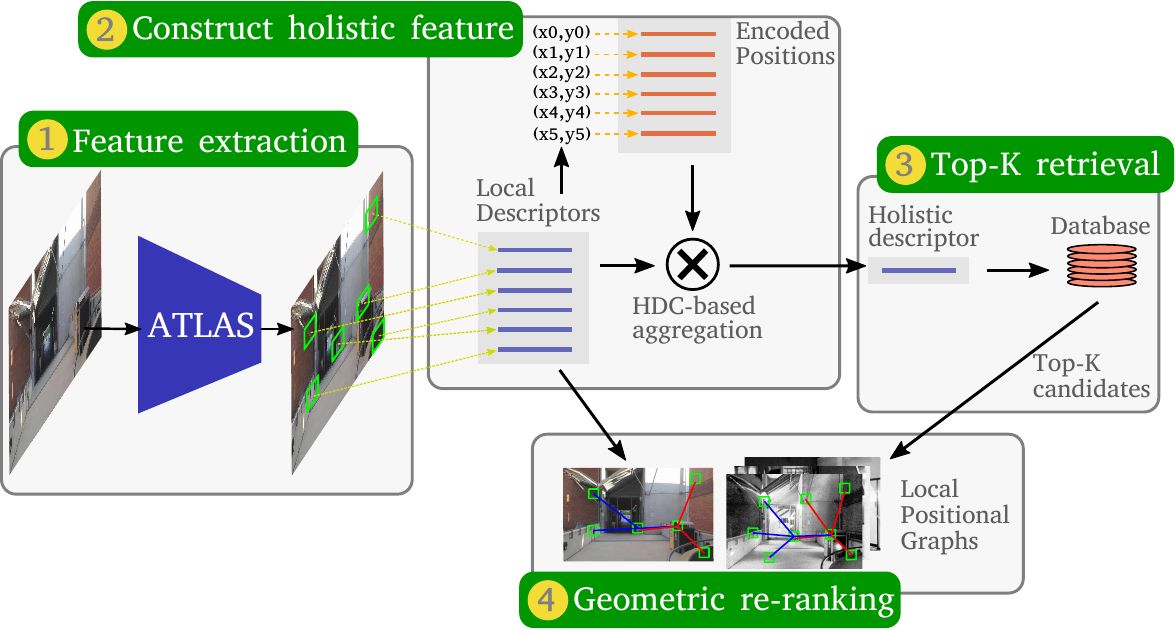}
          
        \caption{An overview of the hierarchical VPR approach proposed in this paper: The pipeline first extracts local ATLAS image descriptors aggregated into a holistic image descriptor with hyperdimensional computing (HDC). These descriptors efficiently retrieve the top-K matching candidates before a final re-ranking with the local descriptors and the proposed Local Positional Graphs.}
  		\label{fig:fig_Overall}
        \vspace{-0.4cm}
	\end{figure}
However, existing best-performing learning-based attentive local feature methods require large-scaled~\cite{DELF} or expensively labeled~\cite{D2Net} training datasets. 

Geometric context among local features in an image significantly enhances the robustness and performance of VPR.
Existing methods addressing lightweight image-wise geometric context~\cite{Patch-NetVLAD}~\cite{Star-Hough} are easy to implement but may have limited enhancement for the VPR performance. 
Others use learning-based methods to learn patch-wise or regional local feature geometric context representation\cite{lightglue}\cite{sun2021loftr}, which are computationally expensive for inference. 

Despite the advantages, local feature-based VPR is computationally expensive, hindering its usage for real-world large-scale VPR. 
Hierarchical VPR pipelines reduce the computational effort by retrieving the top-K candidates in the database with holistic features, then re-ranks the candidates by local features. However, the existing hierarchical VPR pipelines extract local features and holistic features with limited performance~\cite{Patch-NetVLAD} or have lower query speed~\cite{DELG}~\cite{TransVPR}~\cite{Patch-NetVLAD}. 

To address the above-mentioned limitations, this letter proposes the runtime and data-efficient hierarchical VPR pipeline illustrated in~\cref{fig:fig_Overall}.
First, inspired by the attention module of DELF~\cite{DELF}, we extend our local feature approach LocalSPED-SoftMP~\cite{yuan2021softmp} to the attentive local feature pipeline ATLAS.
Compared to our previous work, ATLAS significantly increases the local feature performance in VPR with the same small (24K images) training dataset.
Despite the $7\times$ smaller training set, ATLAS shows a better pairwise mutual matching performance than DELF in our experiments. 
We conducted a detailed ablation study to address the differences between ATLAS and DELF. We found that the use of softmax normalization of the attention scores during training and the use of non-maximum suppression for local feature detection contribute significantly to ATLAS' high performance. 
For a second contribution, we propose a lightweight training-free algorithm coined Local Positional Graph (LPG) to incorporate the local feature patch-wise geometric context for VPR. 
The proposed LPG algorithm shows advantages over several existing image-wise geometric context methods and significantly extends VPR performance for different local feature pipelines, especially for DELF.  
Third, we present a hierarchical VPR pipeline, that uses the same local features (like ATLAS or DELF) first in a hyperdimensional computing (HDC) based holistic descriptor~\cite{HDC_CVPR} for fast candidate selection and then again for candidate re-ranking using the proposed LPG. Hir-ATLAS and Hir-Delf, the combinations with ATLAS and DELF, run up to 14$\times$  faster than ATLAS and DELF but with only a 1.2\% performance drop. We compare Hir-ATLAS and Hir-DELF with the hierarchical pipeline Patch-NetVLAD~\cite{Patch-NetVLAD}, demonstrating advantages in VPR accuracy (+15\%), feature comparison speed (54$\times$), and feature storage occupancy (55$\times$). 

	\begin{figure*}[t]    
  		\centering
  		\includegraphics[width=0.95\linewidth]{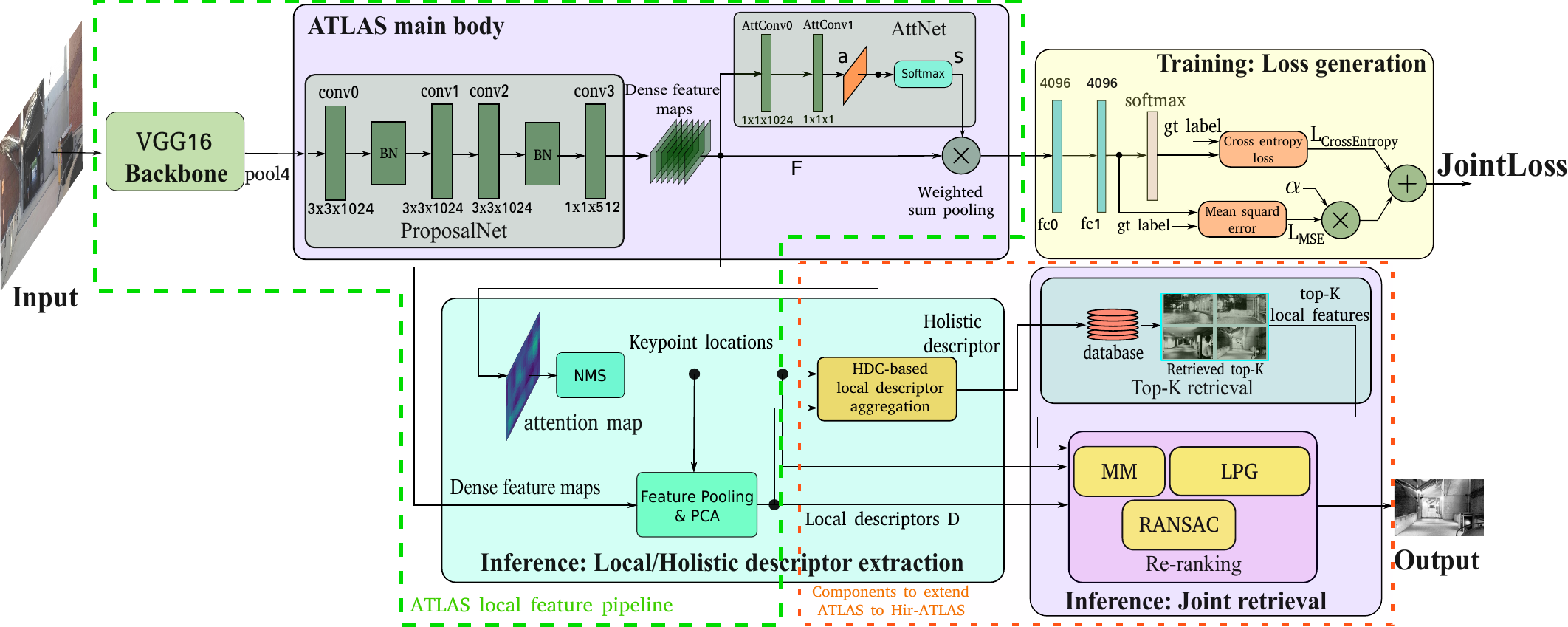}
  		\caption{An overview of the proposed hierarchical pipeline. The green dashed line covered area contains the ATLAS local feature pipeline components. In contrast, the orange dashed line covered area contains the components to extend ATLAS to Hir-ATLAS. Note that the top right yellow block is only used for ATLAS training, whereas the bottom part is exclusively used for inference, i.e., the actual application for VPR. 
  		}  		
  		\label{fig:fig_LocalSPED-AttSoftMP}
        \vspace{-0.2cm}
	\end{figure*}

\section{Related work}
Visual place recognition (VPR) is an active research topic. The recent paper~\cite{Schubert2023} from 2023 provides a detailed introduction to VPR and an overview of relevant publications.
Deep learning-based convolutional neural networks (CNNs) have contributed significantly to the performance improvement of VPR~\cite{VPR-Survey-deep-learning}. 
Especially for the creation of image descriptors, CNNs outperform handcrafted methods under severe appearance changes~\cite{EDGEBOX_CNN}\cite{OLOS}\cite{D2Net}\cite{DELF}\cite{Patch-NetVLAD}.

We can distinguish holistic (global) and local CNN descriptors. Both typically use the feature maps from an intermediate layer in a CNN as descriptors~\cite{HolisticFeatureTranslation}\cite{OLOS}\cite{EDGEBOX_CNN}\cite{Neubert15}\cite{khaliq2019holistic}. 
In earlier work, the regions of interest in an image for local feature-based methods were found either by handcrafted methods~\cite{EDGEBOX_CNN}\cite{Neubert15} or by detecting high activations in a pre-trained CNN~\cite{OLOS}\cite{khaliq2019holistic}. 
However, the performance of these methods was limited because they were not specifically trained for VPR.  
To overcome this limitation, methods specifically trained for VPR began to emerge. In~\cite{SPED}\cite{AttentionHolistic}, a CNN-based holistic descriptor is trained for VPR with images collected from multiple webcams worldwide over a long period. 
To overcome the high memory occupation of the extracted local descriptors for the images~\cite{Patch-NetVLAD}\cite{CTU-ICRA-Paper}, the authors propose attention methods to only detect and describe points of high interest~\cite{SuperPoint}\cite{D2Net}\cite{DELF}.
In SuperPoint~\cite{SuperPoint}, the author train a pixel-level local feature pipeline on a synthetic dataset and warped real-world images using homographic adaptation to allow a self-supervised generation of known point correspondences between image pairs.  
To train D2-Net~\cite{D2Net}, the author uses structure from motion (SfM) to identify the ground truth point correspondences between images and subsequently use them to train a neural network for local feature extraction.
Instead of using known point correspondences for training, DELF~\cite{DELF} uses a (weighted) linear combination of densely extracted local descriptors with a subsequent place classification to learn to detect and describe relevant local features.
Based on \cite{DELF} and \cite{D2Net}, we proposed in our work~\cite{LocalSPED}\cite{yuan2021softmp} novel attentional pooling layers for a local feature detection and description pipeline.
Similar to the training of DELF, we can avoid the creation of point correspondences between image pairs by treating the training as a place classification task with known image correspondences.
Our pipeline can achieve a high local descriptor performance using only a small training dataset.

Besides the local descriptor, the geometric context among the local features can also contribute to improve the performance of VPR~\cite{Graph-Kernels}\cite{Graph3-point}\cite{luo2019contextdesc}\cite{lightglue}\cite{sun2021loftr}. 
However, the existing methods often use complex, time-consuming algorithms such as probabilistic inference~\cite{Graph-Kernels} or random walk~\cite{X-view} for graph representation and comparison or introduce extra neural networks like a graph neural network~\cite{lightglue} or a transformer~\cite{sun2021loftr} for geometric context feature extraction. 
However, there are also lightweight algorithms addressing the local feature geometric context. 
In~\cite{Patch-NetVLAD}, the author proposes the rapid spatial scoring method, which calculates the image similarity with the spatial displace of the mutually matched local features.
In~\cite{Star-Hough}, the proposed Star-Hough algorithm constructs a star-shaped graph of local features in an image. 
However, both methods only address the image-wise geometric context among the local features, so they do not exploit patch-wise geometric details and are sensitive to viewpoint changes.  
In~\cite{CTU-ICRA-Paper}, the author only constructs the local spatial context between the dense extracted local image features in the nearby $3{\times}3$-grid.
Inspired by the above methods, we propose a novel lightweight graph method addressing patch-wise geometric context for image comparison.

Hierarchical VPR pipelines combine the advantages of holistic and local descriptors: Using the holistic descriptors to select the top-K candidates in the database for a query before re-ranking the candidates with the local features~\cite{DELG}\cite{Patch-NetVLAD}.  
However, most holistic descriptor-based methods cannot extract local features~\cite{HolisticFeatureTranslation}\cite{OLOS}\cite{SPED}\cite{ali2023mixvpr}\cite{AttentionHolistic}\cite{NetVLAD}. In contrast, local descriptor methods do not provide holistic image representations~\cite{D2Net}\cite{DELF}.
Using different holistic and local features slows down the query speed.
One versatile approach to aggregate local features in a global descriptor is hyperdimensional computing (HDC)~\cite{HDV-VSA}. HDC-DELF bundles a set of local descriptors with their position information into a single holistic descriptor, allowing the reuse of local descriptors for the holistic descriptors.
In this letter, we use the local feature aggregation of~\cite{HDC_CVPR} as an add-on module to generate holistic descriptors directly from our local descriptors to formulate a hierarchical VPR pipeline.

Despite their hierarchical structure, existing methods are often slow in the re-ranking stage~\cite{DELG}~\cite{Patch-NetVLAD}~\cite{TransVPR} due to a large number of local features extracted per image. The proposed ATLAS pipeline extracts a condensed set of local features per image, significantly reducing the computing time for the re-ranking stage.

\section{Algorithmic description}\label{sec:algo}

This section provides details of the VPR pipeline outlined in~\cref{fig:fig_Overall} and more deteiledly illustrated in~\cref{fig:fig_LocalSPED-AttSoftMP}. We will first present the local feature extraction method ATLAS in~\cref{sec:algo_atlas_attention}, followed by a description of the hierarchical HDC-based local-holistic VPR approach in~\cref{sec:algo_hierarchy}, and finally the details of the proposed Local Positional Graph (LPG) approach in~\cref{sec:algo_lpg}.

\subsection{A detailed overview of ATLAS and its attention mechanism}\label{sec:algo_atlas_attention}
The overall architecture of the proposed hierarchical pipeline is shown in \cref{fig:fig_LocalSPED-AttSoftMP}.
It comprises five main components: \textit{backbone}, \textit{ATLAS main body}, \textit{loss generation}, \textit{local/holistic descriptor extraction}, and \textit{joint retrieval}. 
In the ATLAS local feature pipeline (green dashed line covered area), the \textit{backbone} extracts the dense raw descriptors from the input image using the pool4-layer of VGG16~\cite{VGG16} trained on ImageNet~\cite{imagenet}.
Subsequently, these descriptors are further processed by the \textit{ATLAS main body}, which is composed of the two trainable CNN networks \textit{ProposalNet} and \textit{AttNet} (Attention Net).
ProposalNet transforms the raw descriptors from the backbone into the tensor $F\in\mathbb{R}^{H\times W\times C}$, which contains $H{\times}W$ VPR-specific dense local descriptors of dimensionality $C$.
In ATLAS, we introduce AttNet into our pipeline, which is a two-layer CNN that extracts attention scores $a\in\mathbb{R}^{H\times W}$ from $F$ to identify relevant, robust local features for VPR. The $softmax$-normalization is applied to all elements $a_{yx}\in a$ with
\vspace{-0.3cm}
\begin{equation}
    \forall y,x: \ \ s_{yx} =\frac{e^{a_{yx}}}{\sum_{i=1}^{H}\sum_{j=1}^{W} e^{a_{ij}}}
    \label{eq:softmax}
    \vspace{-0.2cm}
\end{equation}
to obtain normalized attention scores $s\in\mathbb{R}^{H\times W}$.
The final step of the ATLAS main body is the creation of a global representation $I\in\mathbb{R}^C$ of the input image using a weighted sum of all local descriptors in $F$:
\vspace{-0.3cm}
\begin{equation}
    \label{eq:img_global_rep}
    \forall c: \ \ I_c = \sum_{y=1}^{H}\sum_{x=1}^{W}s_{yx}\cdot F_{yxc} \ \ .
    \vspace{-0.2cm}
\end{equation}

For \textbf{training}, $I$ is passed to the \textit{loss generation} module (cf.~\cref{fig:fig_LocalSPED-AttSoftMP}) to obtain the loss of the whole pipeline.
Here, the training of ATLAS is formulated as a place classification task: $I$ serves as input to two fully-connected layers (fc) that try to predict the correct place of the input image from $N$ places.
For the comparison of the predicted place and the ground truth, a combination of mean squared error $L_{MSE}$ and cross entropy $L_{CrossEntropy}$ with
\vspace{-0.3cm}
\begin{equation}
    \label{eq:loss}
    JointLoss =L_{CrossEntropy} + \alpha\cdot L_{MSE}
\end{equation}
is used as proposed in~\cite{yuan2021softmp}.
The constant $\alpha$ weights the mean squared error.
Both losses are defined with
\begin{gather}
    \label{eq:loss_entropy}
    L_{CrossEntropy} = -\sum_{i=1}^{N}gt_i\cdot log(softmax(v_i)) \\
    \label{eq:loss_mse}
    L_{MSE} = \frac{1}{N}\sum_{i=1}^{N}(gt_i-v_i)^2 \ \ .
\end{gather}
The vectors $gt\in\mathbb{B}^N$ and $v\in\mathbb{R}^N$ represent the one-hot encoded ground truth place label of the input image and the predictions from the fully-connected layers (output of $fc1$ in \cref{fig:fig_LocalSPED-AttSoftMP}).

During \textbf{inference}, the dense local descriptors~$F$ and the unnormalized attention scores~$a$ from the ATLAS main body are fed into the module for \textit{local/holistic descriptor extraction} (cf.~\cref{fig:fig_LocalSPED-AttSoftMP}, bottom).
Here, relevant local features with high attention scores are detected by a non-maximum suppression (NMS) with a $3{\times}3$ sliding window.
To create the final set of local descriptors $D$, the descriptors in $F$ are first pooled within a $d{\times}d$-window around the found maxima in $a$ before compressing the flattened patch features with principal component analysis (PCA) from dimensionality $d\cdot d\cdot C$ to a desired local descriptor length $d_{loc}$.
Note that the PCA was previously learned on the training dataset.
\cref{sec:exp_ablation} will provide an extensive ablation study to evaluate the design decisions.

\subsection{Extending ATLAS for hierarchical VPR (Hir-ATLAS)}\label{sec:algo_hierarchy}
We extend ATLAS to Hir-ATLAS for hierarchical VPR by incorporating the local descriptor aggregation method of~\cite{HDC_CVPR} into the module \textit{local/holistic descriptor extraction} (cf.~\cref{fig:fig_LocalSPED-AttSoftMP}). The method uses hyperdimensional computing (HDC) to bundle the set $D$ of local descriptors with their image positions into a single holistic descriptor (coined HDC-ATLAS).
The \textit{joint retrieval} module (cf.~\cref{fig:fig_LocalSPED-AttSoftMP}) can then perform the actual hierarchical VPR: In the first step, it uses the holistic descriptors to retrieve the top-K candidates from the database with the highest cosine similarity.
The K candidates are re-ranked using the local descriptors in the second step.
For re-ranking, we use three different approaches:
\begin{enumerate*}
    \item mutual matching (\textit{MM}) of the local descriptors~\cite{EDGEBOX_CNN},
    \item estimation of the fundamental matrix between two images using \textit{RANSAC}~\cite{Hartley2003}, and
    \item our proposed local positional graph (\textit{LPG}) addressing the local feature patch-wise geometric context (\cref{sec:algo_lpg}).
\end{enumerate*}

The three methods re-rank the image similarities $S_{db,q}$ between the query image and top-K candidates from the database with
\begin{equation}
    \label{eq:im_sim}
    S_{db, q} = \frac{1}{\sqrt{ |D_{db}| \cdot |D_q|}} \sum_{\forall i,j} w_{ij} \cdot \cos( D_{db}^i, D_q^j) \ \ .
\end{equation} 
$D_{db}{=}\{D_{db}^i\}$ and $D_q{=}\{D_q^j\}$ are the sets of local descriptors from the $db$-th or $q$-th database or query image.
$\cos(\cdot)$ is the cosine similarity between both input vectors.
The weighting factor $w_{ij}$ is set by the re-ranking method:
For MM, $w_{ij}=1$ if $\{D_{db}^i, D_q^j\}$ are mutual matches, otherwise $w_{ij}=0$.
For RANSAC, $w_{ij}=1$ if the positions of the matched local features $\{i,j\}$ are inliers, otherwise $w_{ij}=0$.
How $w_{ij}$ is set for LPG is explained in the following.

\begin{table*}[t]
\centering
\caption{
mean AUC performances on different datasets of the exhaustive pairwise image comparison with local descriptors (left). Comparison of the LPG-related method (middle). Ablation study for ATLAS' local descriptor on MM (right). LSPD represents the method LocalSPED-SoftMP~\cite{yuan2021softmp}. RSS represents the Rapid spatial scoring of~\cite{Patch-NetVLAD}} %
\label{TB-ssf}

\resizebox{\textwidth}{!}{%
\begin{tabular}{|l|ccc|ccc||cc|cc|cc|cc||ccccc|}
\hline
\multirow{3}{*}{Dataset}  & 
\multicolumn{6}{c||}{\textbf{Local descriptor performance}} & 
\multicolumn{8}{c||}{\textbf{Methods compared to LPG}} &
\multicolumn{5}{c|}{\textbf{Ablation study: ATLAS' local descriptor + MM}}\\
\cline{2-20}
& 
\multicolumn{3}{c|}{\textbf{MM} }& 
\multicolumn{3}{c||}{\textbf{LPG(ours)} } & 
\multicolumn{2}{c|}{\textbf{RANSAC} }& 
\multicolumn{2}{c|}{\textbf{Star-Hough~\cite{Star-Hough}} } &
\multicolumn{2}{c|}{\textbf{POS~\cite{HDC_CVPR}} } & 
\multicolumn{2}{c||}{\textbf{RSS~\cite{Patch-NetVLAD}} } & 
\multirow{2}{*}{\makecell{orig.}} &
\multirow{2}{*}{\makecell{w/o\\softmax}} & 
\multirow{2}{*}{\makecell{w/o\\MSE}} & 
\multirow{2}{*}{\makecell{w/o\\NMS}} & 
\multirow{2}{*}{\makecell{w/o\\ProposalNet}}
\\ 
\cline{2-9}
\cline{10-15}
 & DELF\cite{DELF} & LSPD\cite{yuan2021softmp} & ATLAS(ours)& DELF  & LSPD & ATLAS  & DELF  & ATLAS & DELF &ATLAS & DELF& ATLAS & DELF& ATLAS &&&&&  \\

\hline
GPW           &                   0.58  & 0.75  & \textbf{0.85}            & 0.87 & 0.85 & \textbf{0.92}              &0.81 & 0.90            &0.84 &0.89 &0.83 &0.84 &0.65 &0.79 & 0.85  & 0.08 & 0.82  & 0.76 &0.76\\

Oxford        &                   0.60  & 0.57  & \textbf{0.69}            & \textbf{0.84} & 0.70 & 0.79              &0.73 & 0.76            &0.79 &0.82 &0.79 &0.76 &0.61 &0.60 & 0.69  & 0.12 & 0.64  & 0.43 &0.59\\

SFU           &                   0.69  & 0.69  &\textbf{0.75}             & \textbf{0.86} & 0.79 & 0.85              &0.78 & 0.80            &0.87 &0.83 &0.82 &0.81 &0.69 &0.68 & 0.75  & 0.07 & 0.76  & 0.79 &0.53\\
 
CMU           &                   0.72  & 0.76  & \textbf{0.77}            & \textbf{0.80} & 0.78 & 0.79              &0.77 & 0.78            &0.79 &0.77 &0.77 &0.77 &0.55 &0.69 & 0.77  & 0.10 & 0.77  & 0.63 &0.71\\

Nordland      &                   0.50  & 0.68  & \textbf{0.75}            & 0.78 & 0.80 & \textbf{0.87}              &0.68 & 0.81            &0.76 &0.84 &0.73 &0.82 &0.51 &0.70 & 0.75  & 0.08 & 0.77  & 0.52 &0.57\\ 

StLucia       &                   0.47  & 0.47  & \textbf{0.50}            & \textbf{0.69} & 0.55 & 0.62              &0.54 & 0.53            &0.67 &0.55 &0.59 &0.55 &0.50 &0.44 & 0.50  & 0.06 & 0.50  & 0.33 &0.43\\
\hline
mean          &                   0.60  & 0.65  & \textbf{0.72}            & \textbf{0.81} & 0.74 & \textbf{0.81}     &0.72 & 0.76            &0.79 &0.79  &0.76 &0.76 &0.58 &0.65 & 0.72  & 0.08 & 0.70  & 0.56 &0.60\\ 
\hline
\end{tabular}
}
\vspace{-0.6cm}
\end{table*}

\begin{table}[t]
\centering
\caption{AUC of ATLAS and DELF local features in GPW day-left vs. night-right sequence with different LPG hyper-parameter combinations of $h\times h$ and $\sigma$. Best combination for both ATLAS and DELF is bold.}
\label{TB-ATLAS-LPG-combine-parameter-Study}

\resizebox{\linewidth}{!}{%
\begin{tabular}{|l|cc|cc|cc|cc|cc|}
\hline  
\scriptsize h$\times$h & 
\multicolumn{2}{c|}{\scriptsize 10$\times$10} & 
\multicolumn{2}{c|}{\scriptsize 20$\times$20} & 

\multicolumn{2}{c|}{\scriptsize 40$\times$40} &

\multicolumn{2}{c|}{\scriptsize 60$\times$60} &

\multicolumn{2}{c|}{\scriptsize 80$\times$80}\\
\hline
Pipeline & \scriptsize ATLAS & \scriptsize DELF & \scriptsize ATLAS & \scriptsize DELF & \scriptsize ATLAS & \scriptsize DELF & \scriptsize ATLAS & \scriptsize DELF & \scriptsize ATLAS & \scriptsize DELF\\
\hline
\scriptsize   $\sigma = 0.5$       &0.77 &0.35      &0.85 &0.55        &0.87 &0.66         &0.88 &0.74          &0.88 &0.77 \\
\scriptsize   $\sigma = 1.0$       &0.83 &0.46      &0.87 &0.66        &0.88 &0.75 &\textbf{0.88} &\textbf{0.78} &0.87 &0.78 \\
\scriptsize   $\sigma = 2.0$       &0.83 &0.50      &0.86 &0.66        &0.87 &0.73         &0.86 &0.75          &0.85 &0.74 \\
\scriptsize   $\sigma = 3.0$       &0.83 &0.48      &0.85 &0.62        &0.85 &0.69         &0.85 &0.72          &0.83 &0.71 \\
\hline

\end{tabular}
}
\vspace{-0.3cm}
\end{table} 

\begin{figure}[t]   
  \centering 
  \includegraphics[width=0.95\linewidth]{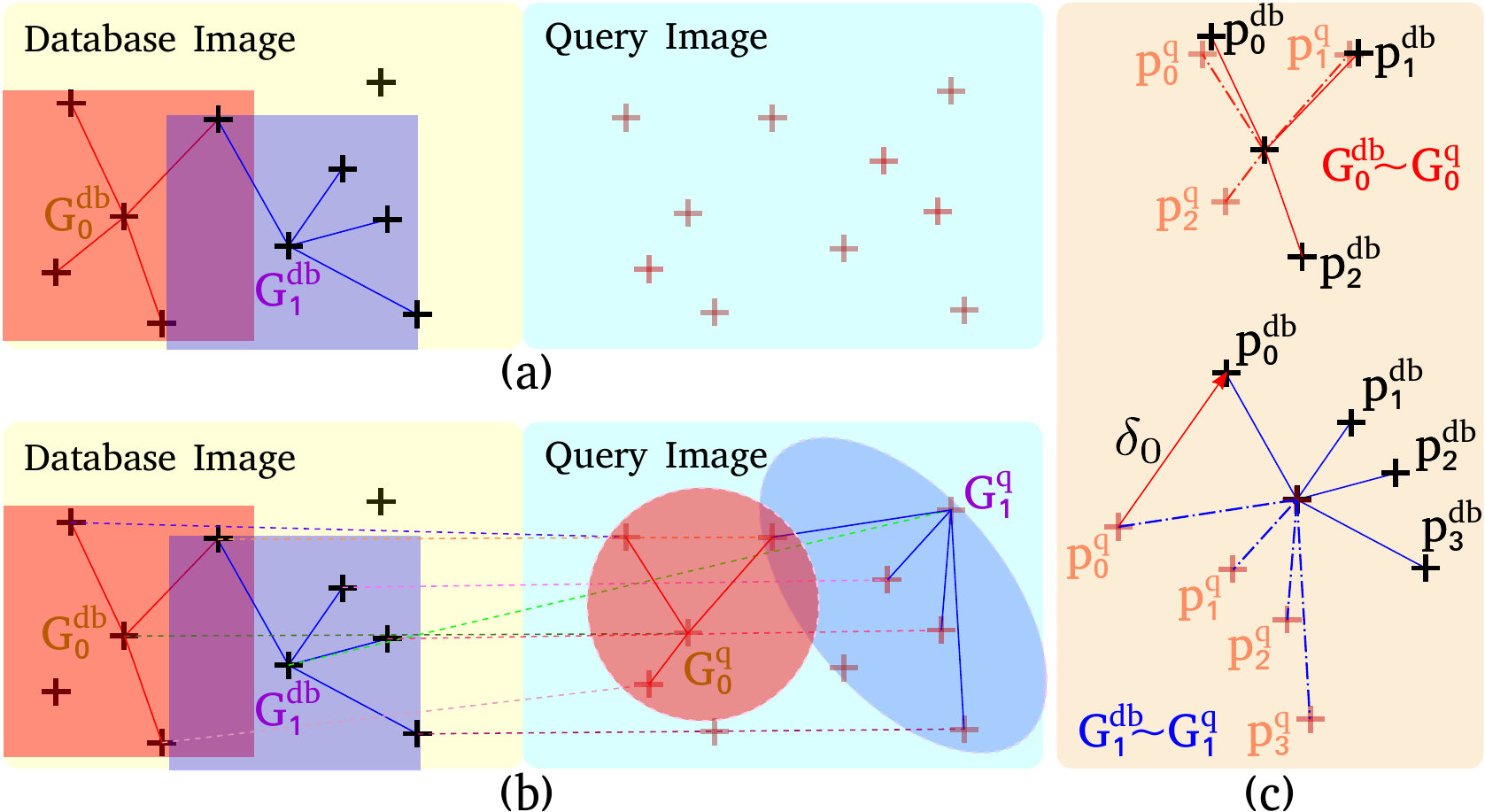}

  \caption{Visualization of the Local Positional Graph (LPG). The crosses represent the local features in the images. (a) Creating a star-shaped graph for each local feature in the database image: Two graphs $G_0^{db}$ and $G_1^{db}$ are created for two local features. (b) Creation of corresponding graphs in the query image: The local features in the query image that are mutually matched to the root nodes in $G_0^{db}$ and $G_1^{db}$ serve as root nodes in the graphs $G_0^q$ and $G_1^q$. Features in the query image that are mutually matched with the leaf nodes of $G_0^{db}$ and $G_1^{db}$ are leaf nodes in $G_0^q$ and $G_1^q$. Unmatched leaves are discarded for $G_0^{db}$ (bottom left node in $G_0^{db}$). (c) Graph comparison: The node positions in each graph are translated with the position of their root nodes to overlay the corresponding graphs. The displacement vectors $\delta_k$ are determined for all corresponding leaf nodes.}
\label{fig:fig_LPG}
\vspace{-0.2cm}
\end{figure}

\subsection{The local positional graph (LPG) for re-ranking}\label{sec:algo_lpg}

To overcome the limitations of MM, which does not consider the geometric context among local features, and of RANSAC, which is computationally very inefficient, we propose the local positional graph (LPG) for evaluating the similarity of two images.
LPG is a lightweight training-free graph-based approach that exploits the patch-wise local feature geometric context to enhance the local feature performance in VPR.
The LPG algorithm is composed of the three steps
\begin{enumerate*}
    \item graph creation,
    \item graph comparison, and
    \item image similarity evaluation.
\end{enumerate*}

\subsubsection{Graph creation}
For a database image with $N$ local features, LPG first creates $N$ star-shaped graphs $G_n^{db}$ for each of the local features, where $n$ indexes the graph constructed for local feature $n$.
As shown in \cref{fig:fig_LPG}~(a), the local feature $n$ in $G_n^{db}$ is the root node, while all surrounding local features within a rectangular window of size $h{\times}h$ (in local feature positional space) are leaf nodes.
After a mutual matching of the local features between a database image and a query image, the corresponding star-shaped graphs $G_n^q$ for the query image can be constructed directly based on $G_n^{db}$, as shown in \cref{fig:fig_LPG}~(b):
The root node of $G_n^q$ is the query image's local feature that mutually matches with the root node of $G_n^{db}$.
The leaf nodes in $G_n^q$ are the query image features that mutually match with the leaf nodes of $G_n^{db}$; leaf nodes in $G_n^{db}$ without a feature match in the query image are ignored in the successive graph comparison step.
The outcome of the graph creation is a set of graph pairs $\{G_n^{db}\sim G_n^q\}$.
\subsubsection{Graph comparison}
For the graph comparison of a pair $\{G_n^{db}\sim G_n^q\}$, we first overlay their root nodes, as shown in \cref{fig:fig_LPG}~(c).
Subsequently, a displacement vector~$\delta_k$ is computed with
\vspace{-0.1cm}
\begin{equation}
    \delta_k = p_k^{db} - p_k^q
\end{equation}
for each mutually matching leaf node pair with positions $p_k^{db}$ and $p_k^{q}$. %
Here, $k\in[1..K]$ is the index of the $K$ matched leaf nodes in the overlayed graph pair $\{G_n^{db}\sim G_n^q\}$.
Next, the squared $\ell2$-norm of each $\delta_k$ is mapped to displacement scores in the range $[0..1]$ with an unnormalized Gaussian
\begin{equation}
    \label{eq:gaussian}
    G(\delta_{k})=exp(-{\frac{\|\delta_{k}\|_2^2}{2\sigma^2}}) \ \ .
\end{equation}
\subsubsection{Image similarity evaluation}
In the final step, we average $G(\delta_{k})$ over all $K$ matched leaf nodes of a graph pair $\{G_n^{db}\sim G_n^q\}$ with
\begin{equation}
    \label{eq:LPG_w}
    w_{ij}=\frac{1}{K}\sum_{k=1}^K G(\delta_{k})
\end{equation}
to calculate the patch-wise geometric context. %
Here, $i$ and $j$ are the indices of the root node features in the database and query images, respectively.
After the comparison of all graph pairs, all $w_{ij}$ can be used to compute the final similarity $S_{db,q}$ of the image pair with \cref{eq:im_sim}.
Note that $w_{ij}=0$ if the $i$-th and $j$-th local features are not mutually matched.

The most time-consuming part of LPG is the graph creation for the database images.
Fortunately, this can be done offline before the application for VPR.
Moreover, the Gaussian function in \cref{eq:gaussian} could be quantized to a reasonable resolution so that its calculation can be converted into a look-up table for efficient computation. The following experiments will demonstrate the low computational overhead of LPG.

\section{Experiments}\label{sec:exp}
The experiments are conducted on a collection of standard visual place recognition datasets. 
First, we compare the local features of ATLAS and DELF and our predecessor local feature pipeline LocalSPED-SoftMP (LSPD), using mutual matching (\textit{MM}) and LPG. We also conduct the hyper-parameter investigation for LPG and compare LPG to three existing related algorithms.
Second, we evaluate the holistic descriptor performance of HDC-ATLAS, HDC-DELF (holistic features aggregate with DELF local features), and the holistic feature used by Patch-NetVLAD.
To compare with Hir-ATLAS, we extend DELF to Hir-DELF using the same methods described in~\cref{sec:algo_hierarchy}.
Next, we evaluate the hierarchical VPR performance of Hir-ATLAS, Hir-DELF, and Patch-NetVLAD on three aspects: VPR performance, query speed, and disk storage occupancy.
Finally, we conduct an ablation study for the ATLAS local feature pipeline.

\subsection{Experimental setup}\label{sec:exp-setup}
\subsubsection{Parameters}
We use all ATLAS local features with their local descriptors. The local features are detected with the non-maximum suppression (NMS) on the attention scores~$a$ using a $3{\times}3$ sliding window. To create the local descriptors $D^i$, we pool the dense local features in $F$ in a $7{\times}7$-window ($d=7$) around a local maximum in $a$ before they are compressed with PCA to dimensionality $d_{loc}=1024$.
For DELF local features, we use the implementation provided by the authors and select the top 200 local descriptors (with the highest scores) of dimensionality 1024.
For LPG, we normalize the feature positions to $[0..100)$. 
We also quantize $\|\delta_{k}\|_2^2$ in this local feature coordinate space to sample the Gaussian function in~\cref{eq:gaussian}. We conduct a hyper-parameter search on \textit{day-left -- night-right} sequence of the GPW dataset to find the best $\sigma$ and $h$ combination for both ATLAS and DELF. As shown in ~\cref{TB-ATLAS-LPG-combine-parameter-Study}, the LPG parameters are set to $\sigma=1.0$ and $h=60$ for maximum performance.
For Patch-NetVLAD, we also use the author's model, which is trained on the mapillary dataset.
For $4096$-dimensional HDC-ATLAS holistic descriptors, we use the HDC-based local descriptor aggregation method from \cite{HDC_CVPR} with $n_x=5$ and $n_y=9$.
We use the holistic descriptor HDC-DELF from \cite{HDC_CVPR} without the feature normalization for the holistic feature of Hir-DELF.

\begin{figure}[t]   
  \centering 
  \includegraphics[width=0.95\linewidth]{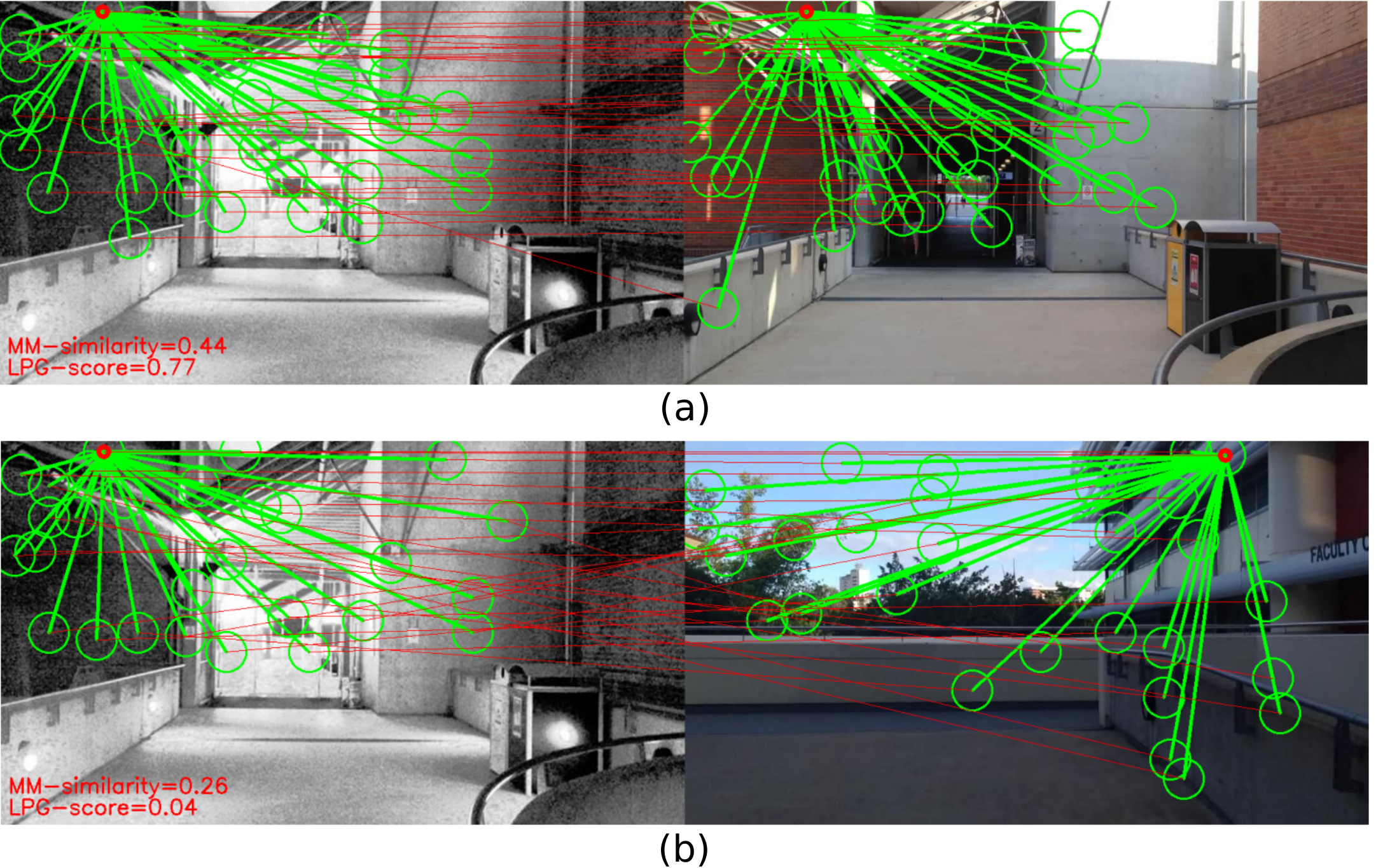}
  \caption{LPG graphs with correctly matched root nodes (a) and incorrectly matched rood nodes (b) using local ATLAS features between database images (left) and query images (right). Red circles show the root nodes. Green circles show leaf nodes with a radius that corresponds to the pixel size in the attention map. Red lines connect matched leaf nodes.}
\label{fig:LPG-demo}
\vspace{-0.3cm}
\end{figure}

\begin{table}[tb]
\centering
\caption{
mean AUC/Recall@100 ($K=100$) performances on different datasets of the holistic descriptors HDC-ATLAS for Hir-ATLAS, HDC-DELF for Hir-DELF, and HOL-PVLAD for Patch-NetVLAD.}
\label{TB-hol-feat-AUC}
\resizebox{0.9\linewidth}{!}{%
\begin{tabular}{|l|cc|cc|cc|}

\hline
\multirow{2}{*}{Dataset}  & \multicolumn{2}{c|}{\textbf{HDC-ATLAS} (ours)} & \multicolumn{2}{c|}{\textbf{HDC-DELF} \cite{HDC_CVPR}} & \multicolumn{2}{c|}{\textbf{HOL-PVLAD} \cite{Patch-NetVLAD}} \\
\cline{2-7}
  &   AUC & R@100  & AUC & R@100 & AUC & R@100 \\
\hline
GPW &         0.52 & 0.98            & 0.63 & \textbf{1.00}                          &  0.44  &  0.97  \\

Oxford &  0.35 &  0.94   & \textbf{0.64} & \textbf{0.95}        & 0.55 &  0.90  \\

SFU &  0.58 & \textbf{0.98}   & \textbf{0.61} & \textbf{0.98}                  & 0.06 & 0.90 \\

CMU &  0.65 & 0.89                 & \textbf{0.68} & 0.89                   & 0.51 & \textbf{0.90} \\

Nordland      & 0.64 & \textbf{0.99}            & \textbf{0.61} & 0.98             & 0.15 & 0.88 \\

StLucia &  0.41 & 0.83       & \textbf{0.42} & \textbf{0.87}              & 0.08 & 0.63 \\

\hline
mean &   0.51 & \textbf{0.94}                                  & \textbf{0.61} & \textbf{0.94}                                        & 0.33 & 0.87 \\

\hline
\end{tabular}
}
\vspace{-0.3cm}
\end{table}

\begin{table*}[t]
\centering
\caption{AUC performances of the hierarchical VPR pipelines. Hir-ATLAS with NVD+RANSAC uses Patch-NetVLAD's holistic descriptor for candidate selection before re-ranking with the ATLAS' local descriptors and RANSAC.}
\label{TB-holpairwise}

\resizebox{0.84\linewidth}{!}{%
\begin{tabular}{|ll|ccc|cccc|cc|}
\hline
\multirow{2}{*}{Dataset} & \multirow{2}{*}{Query - DB}  &
\multicolumn{3}{c|}{\textbf{Hir-DELF top-100}} & 
\multicolumn{4}{c|}{\textbf{Hir-ATLAS top-100} (ours)} &
\multicolumn{2}{c|}{\textbf{Patch-NetVLAD top-100}} \\
\cline{3-11}
 &          & MM  & LPG (ours) & RANSAC & MM & LPG (ours) & RANSAC & NV+RANSAC  & Performance &  Speed \\
\hline
GPW           &     \tiny day-left--day-right              & 0.92 & 0.98 & 0.98    & 0.97 & 0.98 & 0.98                  & 0.99 & \textbf{1.00}            &  \textbf{1.00} \\
              &     \tiny day-right--night-right (night)   & 0.49 & 0.86 & 0.79    & 0.79 & 0.89 & 0.85                  & 0.84 & \textbf{0.92}   & 0.74\\
              &     \tiny day-left--night-right (night)    & 0.26 & 0.78 & 0.66    & 0.75 & \textbf{0.85} & 0.84         & 0.83 & 0.80            & 0.58\\
\hline
Oxford        &     \tiny 14-12-09--15-05-19               & 0.91 & \textbf{0.95} & 0.94    & 0.70 & 0.79 & 0.77                  & 0.82 & 0.67            & 0.61\\  
              &     \tiny 14-12-09--15-08-28               & 0.34 & \textbf{0.75} & 0.56    & 0.48 & 0.62 & 0.56                          & 0.62 & 0.67   & 0.47\\ 
              &     \tiny 14-12-09--14-11-25               & 0.77 & 0.89 & 0.82             & 0.88 & 0.88 & 0.89                  & \textbf{0.91} & 0.65            & 0.60\\ 
              &     \tiny 14-12-09--14-12-16 (night)       & 0.30 & 0.66 & 0.57             & 0.68 & \textbf{0.70} & 0.70                & 0.67 & 0.44            & 0.28\\ 
              &     \tiny 15-05-19--15-02-03               & 0.81 & \textbf{0.96} & 0.94    & 0.68 & 0.93 & 0.92                  & 0.92 & 0.38            & 0.32\\ 
              &     \tiny 15-08-28--14-11-25               & 0.59 & \textbf{0.79} & 0.71    & 0.86 & 0.72 & 0.64                  & 0.67 & 0.51            & 0.38\\ 
\hline
SFU           &     \tiny dry--dusk                        & 0.76 & 0.86 & 0.80     & 0.73 & 0.82 & 0.78                          & 0.78 & \textbf{0.93}   & 0.86\\
              &     \tiny dry--jan                         & 0.58 & 0.85 & 0.76     & 0.77 & 0.86 & 0.82                          & 0.78 & \textbf{0.86}   & 0.72\\
              &     \tiny dry--wet                         & 0.72 & 0.86 & 0.78     & 0.75 & 0.85 & 0.81                          & 0.78 & \textbf{0.92}   & 0.82\\
\hline
CMU           &     \tiny 20110421--20100901               & 0.79 & \textbf{0.83} & 0.80    & 0.81 & \textbf{0.83} & 0.82                 & \textbf{0.83} & 0.65            & 0.54\\
              &     \tiny 20110421--20100915               & 0.74 & 0.78 & 0.77             & 0.78 & \textbf{0.79} & 0.78         & \textbf{0.79} & 0.55            & 0.47\\
              &     \tiny 20110421--20101221               & 0.63 & 0.69 & 0.68             & 0.65 & 0.68 & 0.66                  & \textbf{0.76} & 0.56            & 0.46\\
              &     \tiny 20110421--20110202               & 0.72 & \textbf{0.85} & 0.80    & 0.83 & \textbf{0.85} & 0.83                 & 0.83 & 0.61            & 0.57\\

\hline
Nordland      &     \tiny spring--winter                   & 0.54 & 0.92 & 0.79      & 0.79 & \textbf{0.94} & 0.86                 & 0.79 & 0.85            & 0.80\\ 
              &     \tiny spring--summer                   & 0.45 & 0.77 & 0.70      & 0.80 & 0.88 & 0.84                          & 0.83 & \textbf{0.92}   & 0.87\\
              &     \tiny summer--winter                   & 0.17 & 0.53 & 0.40      & 0.54 & \textbf{0.74} & 0.63                 & 0.57 & 0.68            & 0.55\\ 
              &     \tiny summer--fall                     & 0.83 & 0.93 & 0.92      & 0.93 & 0.94 & 0.94                          & 0.93 & \textbf{0.97}   & 0.96\\
\hline
StLucia       &   \tiny 100909-0845--180809-1545           & 0.34 & \textbf{0.54} & 0.41      & 0.38 & 0.50 & 0.41                  & 0.25 & 0.30   & 0.24\\
              &   \tiny 100909-1000--190809-1410           & 0.53 & \textbf{0.73} & 0.60      & 0.54 & 0.65 & 0.57                  & 0.47 & 0.50   & 0.45\\
              &   \tiny 100909-1210--210809-1210           & 0.69 & \textbf{0.76} & 0.70      & 0.64 & 0.68 & 0.65                  & 0.68 & 0.75   & 0.72\\

\hline
best           &                                           & 0.92 & 0.98 & 0.98              & 0.97 & 0.98 & 0.98                  & 0.99 &  \textbf{1.00}       & \textbf{1.00}\\
worst          &                                           & 0.17 & \textbf{0.53} & 0.40     & 0.38 & 0.50 & 0.41                  & 0.25 & 0.30         & 0.24\\
mean           &                                           & 0.60 & \textbf{0.81} & 0.73     & 0.73 & 0.80 & 0.76                  & 0.75 & 0.70   & 0.61\\ 
\hline
\end{tabular}
}
\vspace{-0.4cm}
\end{table*}

\subsubsection{Training} \label{sec:model_train}
We train ATLAS with the same training procedure and dataset as in our previous work~\cite{yuan2021softmp}.

\subsubsection{Benchmark}
We evaluate all methods on the six different datasets GardensPoint Walking (\textit{GPW})~\cite{GPW-dataset}, \textit{StLucia} (multiple times of day)~\cite{StLucia-dataset}, \textit{Oxford} RobotCar~\cite{Oxford}, \textit{CMU}~\cite{CMU}, \textit{Nordland}~\cite{Nordland}, and \textit{SFU} Mountain~\cite{SFU}.
StLucia, Oxford, and CMU were sampled with one frame per second. All images of each sequence in GPW and SFU are used. We sampled 1,000 images of unique places (without tunnels) for Nordland as in~\cite{HDC_CVPR}.

\subsubsection{Metrics}
For performance evaluation, we use the area under the precision-recall curve (AUC) and Recall@K~\cite{Schubert2023}.

\subsection{Local feature comparison and LPG evaluation}\label{sec:exp_local_lpg}
We first evaluate ATLAS' local descriptor's performance for VPR and compare it with LSPD and DELF.
Therefore, we use the local descriptors to compare all query images with the whole database exhaustively.
We use mutual matching (MM) and LPG to calculate the similarity of each image pair. 
In addition, we compare LPG with three image-wise geometric context-based methods: Star-Hough~\cite{Star-Hough}, POS (as used in~\cite{HDC_CVPR}), and the Rapid Spatial Scoring (RSS) as used in Patch-NetVLAD~\cite{Patch-NetVLAD}.

\cref{TB-ssf} (left) shows the obtained mean AUC performances on different datasets of the three local feature pipelines with MM and LPG. For each dataset, we use the same sequences as in ~\cref{TB-holpairwise}.
ATLAS with MM outperforms both LSPD and DELF on most of the datasets.
This is also reflected in the mean, best- and worst-case performances over all datasets: ATLAS' local descriptor with MM outperforms LSPD by almost $11\%$ and DELF by even $20\%$ in mean AUC.
The LPG algorithm further enhances the local feature performance over MM for all three descriptors: The mean AUC of ATLAS, LSPD, and DELF is improved by approximately $13\%$, $14\%$ and $35\%$. 
Interestingly, with LPG, DELF reaches the same performance as ATLAS.
One possible explanation is that ATLAS already learned to incorporate more geometric context during training than DELF, which then leaves less room for further improvement for ATLAS. On the other hand, DELF can more strongly benefit from the additional geometric context provided by LPG.

The middle of table \cref{TB-ssf} shows the evaluation results of the three local feature image-wise geometric context-based methods Star-Hough, POS, and RSS. 
The numbers are similar for ATLAS and DELF. Only with RSS, there is a noticeable difference in favour of ATLAS.
The results show that LPG provides better performance than the other three context-based methods. 

    \cref{fig:LPG-demo} shows two pairs of LPG graphs between a database image and a query image. The root nodes are actual mutual matches that were matched either correctly (true positive) or incorrectly (false positive). Both examples demonstrate how LPG contributes to a better performance: In the correct example (\cref{fig:LPG-demo}a), LPG further increases the similarity between the root nodes using the leaf nodes, while in the incorrect example (\cref{fig:LPG-demo}b), LPG further decreases the similarity of the root nodes using the leaf nodes.

\subsection{Ablation study of ATLAS' local descriptor}\label{sec:exp_ablation}
In this section, we conduct an ablation study to find out the critical components of the local descriptor pipeline ATLAS, which are different from the local descriptor DELF~\cite{DELF} and contribute to the high performance of ATLAS.
The four key components that only appear in ATLAS but not in DELF are (cf.~\cref{fig:fig_LocalSPED-AttSoftMP})
\begin{enumerate}
    \item a \textit{softmax} normalization for the attention scores from the attention network AttNet (cf.~\cref{eq:softmax}),\label{comp1}
    \item the mean squared error $L_{MSE}$ (\textit{MSE}) that is used in addition to the cross entropy loss during training (cf.~\cref{eq:loss}),\label{comp2}
    \item a non-maximum suppression (\textit{NMS}) of the attention scores to select relevant local features instead of using the $K$ highest scored local features as in DELF, and\label{comp3}
    \item \textit{ProposalNet} that post-processes the dense local features from VGG16. \label{comp4}
\end{enumerate}
For the ablation study, we run four experiments and omit either component~\ref{comp1}, \ref{comp2}, or \ref{comp4} or replace component~\ref{comp3} with a selection of the 200 highest scored local features as in DELF. We omit \ref{comp4} by extracting the local feature from VGG16 pool4 layer output feature maps (the input feature maps for ProposalNet) instead of ProposalNet's output according to the detected keypoint locations.
The modified pipelines addressing \ref{comp1} and \ref{comp2} are retrained with the same training procedure as with ATLAS (cf.~\cref{sec:model_train}).
We then evaluate the local descriptors with the trained modified pipeline and perform an exhaustive pairwise image comparison between the query set and the database using MM as described in \cref{sec:exp-setup}.
\cref{TB-ssf} (right) shows the obtained results with the original local ATLAS descriptor (column orig.) and with the four modified versions.
The performance measurements show that the softmax normalization (component~\ref{comp1}) is essential during training. Without it, the pipeline fails to learn a meaningful local descriptor for VPR, so the performance drops to nearly zero. 
The mean squared error (component~\ref{comp2}) barely contributes to the performance and only yields a $3\%$ performance gain. However, we observed in the experiment that the additional mean squared error term could increase the sparsity of the attention scores after NMS: When trained with the combined loss function (i.e., $\alpha=10$ in \cref{eq:loss}), the pipeline extracts $15\%$ less local descriptors but provides better VPR performance.
Omitting the non-maximum suppression (component~\ref{comp3}) or ProposalNet (component~\ref{comp4}) also clearly decreases the performance of the local ATLAS descriptors by $22\%$ or $17\%$, respectively, indicating both contribute to ATLAS.
The ablation study indicates, the attention score softmax normalization and the joint loss function for training, and the use of NMS for local feature detection jointly contribute to the performance of ATLAS. 

\subsection{Evaluations of Hir-ATLAS, Hir-DELF, and Patch-NetVLAD} \label{hierarchical-VPR}
In this section, we evaluate the performance of Hir-ATLAS for hierarchical VPR as described in \cref{sec:algo} and compare it with Hir-DELF and the state-of-the-art hierarchical approach Patch-NetVLAD~\cite{Patch-NetVLAD}.
All three methods retrieve and re-rank the top 100 image pairs per query.
Patch-NetVLAD is evaluating either in \textit{performance mode} or \textit{speed mode}: Both use NetVLAD~\cite{NetVLAD} based holistic feature (here we name this holistic feature HOL-PVLAD) to select the top-100 candidates. For re-ranking, the \textit{performance mode} builds upon the multi-scale image patch feature and uses RANSAC-based fundamental
matrix estimation for outlier rejection, while the \textit{speed mode} only uses a single-scale patch feature and the RSS algorithm.
We use MM, LPG or RANSAC (cf.~\cref{sec:algo_hierarchy,sec:algo_lpg}) to re-rank either Hir-ATLAS or Hir-DELF.

In the first experiment, we measure the candidate selection performance of the holistic descriptors HDC-ATLAS, HDC-DELF, and HOL-PVLAD used for the three methods Hir-ATLAS, Hir-DELF, and Patch-NetVLAD.
\cref{TB-hol-feat-AUC} shows the corresponding AUC and Recall@100 ($K=100$) values.
On average, HDC-DELF slightly outperforms HDC-ATLAS on both metrics, while HOL-PVLAD performs worse than HDC-DELF and HDC-ATLAS.

Next, we evaluate the VPR performance of the three hierarchical methods.
The obtained results are shown in \cref{TB-holpairwise}.
On average, Hir-ATLAS performs best, outperforms Hir-DELF on MM and RANSAC configurations, and outperforms Patch-NetVLAD in both \textit{speed} and \textit{performance} mode.
Although Hir-DELF uses the best-performing holistic descriptor HDC-DELF for candidate selection, Hir-DELF in LPG configuration takes only a tiny advantage in mean AUC than Hir-ATLAS with LPG.
Compared to Patch-NetVLAD, Hir-ATLAS with LPG achieves approx. $14\%$ to $31\%$ higher mean AUC performance.
Only on the SFU dataset, Patch-NetVLAD is best suited and outperforms all other methods.
We also evaluate Hir-ATLAS using the holistic descriptor HOL-PVLAD for top-100 candidate selection and using ATLAS' local feature with RANSAC for re-ranking (referred to as NV+RANSAC) to allow a comparison of the local descriptors between ATLAS and Patch-NetVLAD.
In this setup, Hir-ATLAS (with NV+RANSAC) still performs $7\%$ better in mean AUC than Patch-NetVLAD in \textit{performance} setup, which implies a better local feature performance of Hir-ATLAS than Patch-NetVLAD for VPR.

We finally compare the quality of the re-ranking between the best-performing pipeline setups of Hir-ATLAS, Hir-DELF, and Patch-NetVLAD.
We, therefore, measure the Recall@K performance with $K=\{1,5,10,20\}$ after re-ranking the top 100 candidates.
\cref{Recall-K} shows the obtained average recall of the entire benchmark datasets:
On average, Hir-DELF with LPG performs best for all $K$ values.
Hir-DELF and Hir-ATLAS share similar Recall@K performance for most of the $k$ values and outperform Patch-NetVLAD in \textit{performance mode}.

\begin{table}[tb]
\centering
\caption{Average Recall@K over all benchmark datasets.}
\label{Recall-K}

\resizebox{0.87\linewidth}{!}{%
\begin{tabular}{|l|c|c|c|c|}
    \hline
    Pipeline                            &      R@1      &      R@5      &     R@10      &     R@20      \\ \hline
    Hir-ATLAS top-100 LPG (ours)       &\textbf{0.55}&    0.68      & \textbf{0.74}      &  0.80 \\
    Hir-ATLAS top-100 RANSAC (ours)         &     0.53      &     0.67      & 0.73      &     0.80      \\ \hline
    Hir-DELF top-100 LPG                &     0.54      &\textbf{0.69}&\textbf{0.74}& \textbf{0.81} \\
    Hir-DELF top-100 RANSAC               &     0.50      &     0.66      &     0.72      &     0.80      \\ \hline
    Patch-NetVLAD top-100 performance   &     0.52      &     0.63      &     0.68      &     0.74      \\ \hline
\end{tabular}
}
\vspace{-0.3cm}
\end{table}

In summary, our pipeline Hir-ATLAS outperforms Patch-NetVLAD in all re-ranking methods and outperforms Hir-DELF in MM and RANSAC. 
The LPG significantly improves the performance of Hir-DELF and reaches the same performance as Hir-ATLAS in hierarchical VPR.

\subsection{Runtime and disk usage}
Besides the AUC value and recall, other essential factors for practical applications are the query speed and disk space to store the image features.
The feature comparison between query image features and the database image features takes the majority of query time. 
Therefore, we measured only the feature comparison time for the query.
We measure the feature comparison time on the entire benchmark with up to 27K queries. We use a computation platform with Intel i7-4790 CPU and NVIDIA RTX4090 GPU. 
\cref{TB-holpairwise-resources} shows the obtained total feature comparison time, average feature comparison latency, relative speed-up reference to PatchNetVLAD in performance mode, feature disk space occupancy to store all the benchmark image features, and the corresponding mean AUC (mAUC) value of the three pipelines with different setups. 
Patch-NetVLAD in \textit{performance} mode takes 1,400GB disk space to store all the benchmark image features, which is  $55{\times}$ larger than Hir-ATLAS and Hir-DELF both use roughly 25.5GB disk space. 
This is because Patch-NetVLAD in performance mode extracts 2,826 4096-dimensional local features per image, while ATLAS and DELF extract, on average, only 200 1024-dimensional local features per image.

Patch-NetVLAD in \textit{speed} mode provides lower VPR performance, requires more memory, and is not as fast as Hir-ATLAS and Hir-DELF in LPG mode.
Compared to Hir-ATLAS and Hir-DELF in RANSAC setup, Patch-NetVLAD in performance mode runs $8\times$ and $4.68\times$ slower.
The best-performing Hir-ATLAS and Hir-DELF setup (Top-100 LPG) runs even $54\times$ and $74\times$ faster than the best-performing Patch-NetVLAD setup (top-100 performance).
The average feature comparison latency of setup Top-100 LPG for Hir-ATLAS and Hir-DELF takes less than 27ms, which can be regarded as real-time for many practical applications. 
For the comparison between three re-ranking methods MM, LPG, and RANSAC in Hir-ATLAS and Hir-DELF, 
LPG is $1.4\times$ slower than MM but provides better performance. LPG also performs better than RANSAC and runs $6.7\times$ and $15.8\times$ faster for Hir-ATLAS and Hir-DELF.
Finally, the results demonstrate that the hierarchical approach is faster than the exhaustive approaches: The hierarchical approach with top-100 candidates runs approximately $14\times$ faster than the exhaustive local descriptor comparison but with negligible performance drops. 
Despite different use cases, we also compared ATLAS+LPG and LightGlue~\cite{lightglue} with local SuperPoint features~\cite{SuperPoint} on the GPW and SFU datasets and found that ATLAS runs 150$\times$ to 400$\times$ faster than LightGlue, is more robust to viewpoint changes, and requires significantly fewer local features.
\begin{table}[tb]
\centering
\caption{Feature comparison time and disk usage of different pipelines with different setups. The speed up column uses Patch-NetVLAD top-100 Performance as a reference.}
\label{TB-holpairwise-resources}

\resizebox{1\linewidth}{!}{%
\begin{tabular}{|l|c|c|c|c|c|}
    \hline
    Setup & 
    Total comparison time & 
    Avg latency &
    Speed up & 
    Disk usage  &     
    mAUC      \\  
    \hline
    Patch-NetVLAD top-100 performance   & $\sim$11.1h   & 1400ms & 1$\times$ & $\sim$1400GB &     0.70      \\
    Hir-ATLAS top-100 RANSAC (ours)         & 4956s     & 180ms  & 8$\times$   & $\sim$25.5GB   & \textbf{0.76}      \\ 
    Hir-DELF top-100 RANSAC               & 8538s       & 310ms  & 4.68$\times$  & $\sim$25.5GB   &     0.73      \\ 
    \hline
    Patch-NetVLAD top-100 speed         &  2594s   & 95ms & 14.7$\times$ & $\sim$419GB &     0.60      \\     
    Hir-ATLAS top-100 LPG (ours)            & 741s        & 27ms & 54$\times$ & $\sim$25.5GB   & 0.80 \\
    Hir-DELF top-100 LPG                  & 542s        & 20ms & 74$\times$ & $\sim$25.5GB   & \textbf{0.81}      \\ \hline

    Hir-ATLAS top-100 MM (ours)             & 526s        & 19ms & 75$\times$  & $\sim$25.5GB   & \textbf{0.73}      \\
    Hir-DELF top-100 MM                   & 388s        & 14ms & 103$\times$ & $\sim$25.5GB    &     0.60      \\ \hline

    ATLAS exhaustive MM (ours)          & 7383s       & 270ms & 5.4$\times$ & $\sim$25.5GB   & \textbf{0.72}      \\
    DELF exhaustive MM                  & 5170s       & 190ms & 7.7$\times$ & $\sim$25.5GB   &     0.60      \\ \hline
\end{tabular}
}
\vspace{-0.4cm}
\end{table}

\section{Conclusion}
In this letter, we propose the local feature pipeline ATLAS which achieves a decent VPR performance based on a small training dataset. 
The proposed LPG algorithm demonstrates a significant VPR performance enhancement for state-of-the-art local feature pipelines over existing spatial context-based methods. 
Using LPG and the HDC method we extend ATLAS to Hir-ATLAS and DELF to Hir-DELF for hierarchical VPR. 
The high performance, low training cost, low latency for query, and small disk occupancy make Hir-ATLAS and Hir-DELF promising for real-world large-scale VPR in changing environments. 

\vspace{-0.3cm}

\bibliographystyle{IEEEtran}
\bibliography{ref}

\end{document}